\documentclass[10pt,twocolumn,letterpaper]{article}

\usepackage[pagenumbers]{cvpr} %

\newcommand{\tocite}[1]{{\textcolor{red}{[TOCITE:]}}}

\usepackage{color, colortbl}

\definecolor{colorfirst}{rgb}{.866,.945, 0.831} %
\definecolor{colorsecond}{rgb}{1, 0.98, 0.83} %
\definecolor{colorthird}{rgb}{0.76, 0.87, 0.92} %

\newcommand{\cellfirst}{\cellcolor{colorfirst}}
\newcommand{\cellsecond}{\cellcolor{colorsecond}}

\renewcommand{\paragraph}[1]{\vspace{0.5em}\noindent\textbf{#1}.}

\usepackage{float}
\usepackage{makecell}

\definecolor{cvprblue}{rgb}{0.21,0.49,0.74}
\usepackage[pagebackref,breaklinks,colorlinks,allcolors=cvprblue]{hyperref}
\usepackage{bm}
\usepackage{multirow}

\def\paperID{12785} %
\def\confName{CVPR}
\def\confYear{2025}

\newcommand\blfootnote[1]{%
  \begingroup
  \renewcommand\thefootnote{}\footnote{#1}%
  \addtocounter{footnote}{-1}%
  \endgroup
}

\title{FreeTimeGS: Free Gaussian Primitives at Anytime and Anywhere\\ for Dynamic Scene Reconstruction}

\author{
    Yifan Wang\textsuperscript{1*} \quad
    Peishan Yang\textsuperscript{1*} \quad
    Zhen Xu\textsuperscript{1*} \quad
    Jiaming Sun\textsuperscript{1} \quad
    Zhanhua Zhang\textsuperscript{2} \quad
    \\[2pt]
    Yong Chen\textsuperscript{2} \quad
    Hujun Bao\textsuperscript{1} \quad
    Sida Peng\textsuperscript{1} \quad
    Xiaowei Zhou\textsuperscript{1$^{\dagger}$} \quad
    \\[5pt]
    $^1$Zhejiang University \qquad
    $^2$Geely Automobile Research Institute
}
\begin{document}

\twocolumn[
    \maketitle
    \vspace{-3em}
    \begin{center}
    \captionsetup{type=figure}

    \centering

    \includegraphics[width=\linewidth]{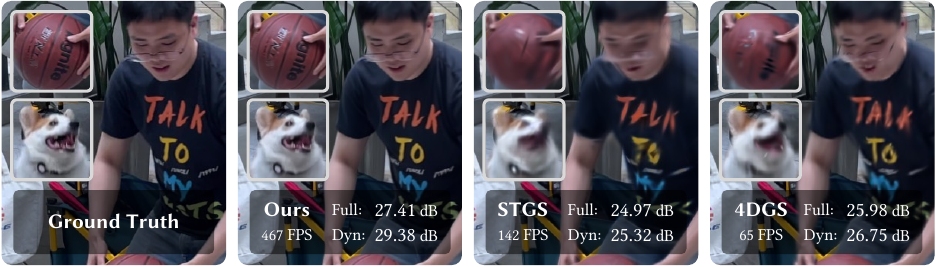}
    \vspace{-8pt}

    \captionof{figure}{%
    \textbf{Photorealistic and real-time rendering of dynamic 3D scenes.} 
    Our method achieves the best performance on challenging dynamic scenes with fast and complex motion.
    Compared with current state-of-the-art methods 4DGS~\cite{yang2023realtime} and STGS~\cite{Li2023SpacetimeGF}, our PSNR is improved by 2.4dB and 1.4dB on the SelfCap dataset. For dynamic regions, our PSNR is improved by 4.1dB and 2.6dB.
    What's more, our method supports real-time rendering at 1080p resolution with a speed of 450 FPS using a single RTX 4090 GPU.
    }
    \label{fig:teaser}
\end{center}

]
\begin{abstract}

This paper addresses the challenge of reconstructing dynamic 3D scenes with complex motions.
Some recent works define 3D Gaussian primitives in the canonical space and use deformation fields to map canonical primitives to observation spaces, achieving real-time dynamic view synthesis.
However, these methods often struggle to handle scenes with complex motions due to the difficulty of optimizing deformation fields.
To overcome this problem, we propose FreeTimeGS, a novel 4D representation that allows Gaussian primitives to appear at arbitrary time and locations.
In contrast to canonical Gaussian primitives, our representation possesses the strong flexibility, thus improving the ability to model dynamic 3D scenes.
In addition, we endow each Gaussian primitive with an motion function, allowing it to move to neighboring regions over time, which reduces the temporal redundancy.
Experiments results on several datasets show that the rendering quality of our method outperforms recent methods by a large margin.
Project page: https://zju3dv.github.io/freetimegs/
\end{abstract}
    
\blfootnote{$^*$Equal contribution. $^\dagger$Corresponding author: Xiaowei Zhou.}

\section{Introduction}
\label{sec:intro}

Dynamic view synthesis aims to produce novel views of a dynamic 3D scene from captured multi-view videos, which has a wide range of applications, such as movie production, video games, and virtual reality. 
Traditional approaches~\cite{hilsmann2020going,casas20144d,collet2015high,newcombe2015dynamicfusion,dou2016fusion4d,orts2016holoportation,yu2018doublefusion} use sequences of textured meshes to represent dynamic 3D scenes.
Such representation requires complicated hardware setups for high-quality reconstruction, thus making traditional approaches limited to controlled environments.
NeRF-based methods~\cite{mildenhall2021nerf,fridovich2023k,li2022neural} have achieved impressive results in dynamic view synthesis by modeling scenes as neural implicit representations.
However, these representations are computationally expensive, leading to slow rendering speed and hindering practical applications.

Recently, a popular paradigm for dynamic view synthesis is to equip 3D Gaussian primitives~\cite{yang2023deformable,wu20234d,Shaw2023SWinGSSW,Guo2024Motionaware3G,Bae2024PerGaussianED,Liang2023GauFReGD,Lu20243DGD,
Labe2024DGDD3,Liu2024MoDGSDG,zhu2024motiongs} with deformation fields to model dynamic scenes.
These methods model the geometry and appearance of the scene based on Gaussian primitives in canonical space and then use MLP networks to model scene motions for deforming the canonical-space scene to observation-space scenes at particular moments.
Although these methods achieve real-time and high-quality rendering performance on scenes with small motions, they often struggle to handle scenes with complex motions.
A plausible reason is that when objects move much in the scene, these methods need to build long-range correspondences between canonical space and observation spaces, which is difficult to be recovered from RGB observations, as discussed in \cite{park2021nerfies, li2020neural}.

In this paper, we propose a novel 4D representation, named FreeTimeGS, for reconstructing dynamic 3D scenes with complex motions.
In contrast to previous methods~\cite{yang2023deformable,wu20234d} that define Gaussian primitives at canonical space only, FreeTimeGS allows Gaussian primitives to appear at arbitrary positions and time steps, possessing the strong flexibility.
In addition, we assign an explicit motion function to each Gaussian primitive, allowing it to move to neighboring regions over time, which facilitates the reuse of Gaussian primitives along the temporal dimension and reduces the representation redundancy.
By endowing Gaussian primitives high degrees of freedom, our representation has two advantages.
First, this significantly improves the ability to model dynamic 3D scenes and the rendering quality, as demonstrated in our experiments.
Second, we only need to model short-range motions between Gaussian primitives and observed scenes, compared with deformation-based methods~\cite{yang2023deformable,wu20234d}.
Therefore, our motion function can be implemented as a linear function, alleviating the ill-posed optimization problem.

During experiments, we find that optimizing FreeTimeGS with only rendering loss tend to get trapped in local minima on fast-moving regions, leading to poor rendering quality.
To tackle this issue, we examine the opacity distribution of Gaussian primitives and discover that a considerable part of it approaches 1.
The result suggests that high opacity of some Gaussian primitives may prevent the gradient from back-propagating to all Gaussian primitives and block the optimization process.
Motivated by this observation, we design a simple regularization strategy to penalize the high opacity of Gaussian primitives in the early stage of optimization, which effectively mitigates the local minima problem and improves our rendering quality.

To validate the effectiveness of our method, we evaluate FreeTimeGS on multiple widely used datasets for multiview dynamic novel view synthesis, including 
Neural3DV~\cite{li2021neural3d} and ENeRF-Outdoor~\cite{lin2022enerf}. 
Our method achieves highest quality on these public datasets compared with existing state-of-the-art methods.
To further evaluate and demonstrate the capability of our method on challenging scenes, we also collect a dataset with much faster and more complex motions compared with Neural3DV~\cite{li2021neural3d}.
Our method achieves best quality on it compared with the SOTA methods by a large margin on both quality and efficiency.

\begin{figure*}[tp]
    \centering
    \vspace{-0.95 cm}
    \includegraphics[width=1.0\linewidth]{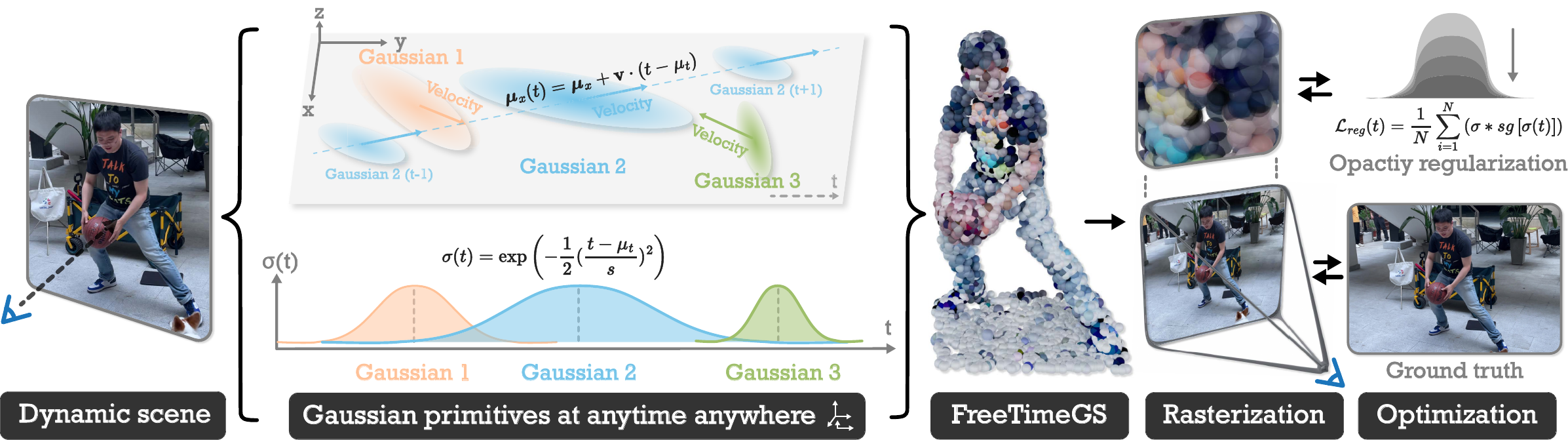}
    \vspace{-0.25 cm}
    \caption{\textbf{Pipeline Overview.} 
    We represent a dynamic scene using Gaussian primitives that can appear anytime anywhere.
    Each Gaussian is assigned with a motion function to to model its movement.
    And its opacity is modulated by the temporal opacity function which control the impact of the Gaussian primitive over time.
    With this 4D representation, we further regularize the Gaussians with a 4D regularization loss and optimize the rasterization result with rendering loss for reconstructing a dynamic 3D scene from multi-view videos.
    }
    \label{fig:allpipeline}
    \vspace{-0.35 cm}
\end{figure*}

\section{Related Work}
\label{sec:related_work}

\paragraph{Dynamic scene reconstruction with RGB-D cameras}
In the early days, dynamic scene reconstruction leverages RGB-D cameras to capture the dynamic geometry and appearance of a scene. Methods pioneered by DynamicFusion~\cite{newcombe2015dynamicfusion} use depth sensors to build a TSDF model in canonical space. Each frame applies non-rigid tracking to create a deformation graph, enabling TSDF fusion in canonical space to integrate reconstruction results across frames. 
Fusion4D~\cite{dou2016fusion4d} and Holoportation~\cite{orts2016holoportation} further refined this system, achieving highly impressive dynamic reconstruction results.
Subsequent works, such as DoubleFusion~\cite{yu2018doublefusion} and BodyFusion~\cite{yu2017bodyfusion}, introduced human body priors, enhancing the robustness of non-rigid tracking. 

The main bottleneck of the aforementioned methods lies in the lack of robustness and inaccuracy of non-rigid tracking. Data-driven approaches, such as DeepDeform~\cite{bozic2020deepdeform} and neural non-rigid tracking, address this by combining pre-trained optical flow networks and using CNNs to predict non-rigid correspondences, achieving promising results. 
However, since these methods cannot optimize visual appearance in an end-to-end manner, they still suffer from issues with robustness and poor visual quality. 
Additionally, they rely on extra depth sensors, which increases the system complexity and overall cost.

\paragraph{NeRF-based dynamic scene reconstruction}
With the rise of NeRF~\cite{mildenhall2021nerf} and differentiable rendering, neural scene representations have gradually become the mainstream approach for dynamic scene reconstruction. Similar to the RGB-D methods discussed above, methods like ~\cite{park2021nerfies, pumarola2021d, park2021hypernerf} build a canonical NeRF model and use an MLP to capture the deformation field from each frame to the canonical model. Another line of research, including NeuralBody~\cite{peng2021neural} and subsequent works~\cite{kwon2021neural,shuai2022multinb,cheng2022generalizable}, utilizes body shape priors, storing per-frame structured latent codes on the body parametric model to model the frame-dependent dynamic appearances and geometries.

Neural3DV~\cite{li2021neural3d} opts to use time-conditioned neural representations to directly model dynamic scenes in the 4D space. 
This approach provides high representational capacity, but modeling dynamics directly in 4D introduces considerable computational costs, leading to long training times and high resource demands
To address efficiency and scalability, hybrid representations such as K-Planes~\cite{fridovich2023k} and HEX-Plane~\cite{cao2023hexplane} combine voxel grids with neural fields, substantially reducing training and inference times. Meanwhile, data-efficient methods~\cite{Takikawa2022VariableBN, tang2022compressible} utilize tensor factorization to enable compact, long-sequence storage.

Despite the impressive progress achieved by NeRF-based methods, challenges in dynamic scene reconstruction remain on slow rendering speeds, poor rendering quality and high storage requirements. In contrast, our proposed approach demonstrates clear advantages in these aspects, making it a more feasible solution for scalable dynamic scene reconstruction.

\paragraph{Gaussian-based dynamic scene reconstruction}
Recently, 3D Gaussian Splatting (3DGS)~\cite{kerbl20233d} has gained popularity as a mainstream approach due to its real-time rendering speed and sharp rendering quality. 
Dy3DGS~\cite{luiten2023dynamic} first adapts 3DGS to dynamic scenes by tracking Gaussian primitives on a frame-by-frame basis.
Similar to RGB-D and NeRF-based methods, ~\cite{yang2023deformable,wu20234d,Shaw2023SWinGSSW,Guo2024Motionaware3G,Bae2024PerGaussianED,Liang2023GauFReGD,Lu20243DGD,
Labe2024DGDD3,Liu2024MoDGSDG,zhu2024motiongs} model the geometry and appearance of the scene in canonical space and then use deformation MLP networks to model scene motions.

Notably, instead of using deformation to model dynamic, 4DGS~\cite{yang2023realtime} and STGS~\cite{Li2023SpacetimeGF} leverage 4D Gaussian primitives to represent dynamic scenes.
However, these method still struggle to handle scenes with complex motions due to their motion representation.
4DGS entangles geometry and velocity, making it difficult to optimize both. Specifically, its spatial scale can be rotated by the velocity vector, converted into temporal scale and vice versa.
What's more, instead of Euclidean space, they optimize the velocity in angular space, where in the case of fast motion, small angular changes can lead to large Euclidean velocity changes, making it difficult to converge.
STGS explicitly models motion using polynomials and angular velocity, but this method has too many parameters and is difficult to optimize in complex motion scenarios, leading to overfitting.
LongVolCap~\cite{xu2024longvolcap} also uses 4D Gaussian primitive representations, but it focuses on modeling long volumetric videos, which is orthogonal to our work and can be further integrated.

\section{Method}

Given multi-view videos that capture a dynamic scene, our goal is to generate novel views of the target scene at arbitrary time.
Our key idea is designing a novel 4D representation, named FreeTimeGS, which leverages Gaussian primitives at any time and location to faithfully represent the content of the dynamic scene.
Figure~\ref{fig:allpipeline} presents the overview of the proposed approach.
In Sec.~\ref{sec:4D_representation}, we first introduce the design details of our 4D representation.
Then, Sec.~\ref{sec:optimization_strategies} describes the optimization strategies to train the 4D representation, which includes 4D regularization, periodic relocation, and 4D initialization.

\subsection{Gaussian primitives at anytime anywhere}
\label{sec:4D_representation}

To represent the content of dynamic 3D scenes, we define Gaussian primitives that can appear at any spatial position and time step.
In addition, our approach assigns a motion function to each Gaussian primitive, allowing it to dynamically adjust its position over time to the neighboring region, thereby enhancing its ability of representing the geometry and appearance of the dynamic scene.
In addition, our approach assigns a motion function to each Gaussian primitive, allowing it to dynamically adjust its position over time to the neighboring region to simulate the physical movement of real-world dynamic objects in the scene, thereby enhancing its ability to represent the evolving geometry and appearance of the dynamic scene.

Specifically, each Gaussian primitive consists of eight learnable parameters: position, time, duration, velocity, scale, orientation, opacity and spherical harmonics coefficients.
To calculate the opacity and color of the Gaussian primitive at any $(\mathbf{x}, t)$, we first move the Gaussian primitive to obtain its actual spatial position $\bm{\mu}_x(t)$ at time $t$ according to its motion function, which is defined as:
\begin{equation}
    \label{eq:position}
    \bm{\mu}_x(t) = \bm{\mu}_x + \mathbf{v} \cdot (t - \mu_t),
\end{equation}
where $\mathbf{v} \in \mathbb{R}^3$ is the velocity of Gaussian primitive, and $\bm{\mu}_x$ and $\mu_t$ are the original position and time of Gaussian primitive, respectively.

Based on the moved Gaussian primitive, we calculate its color at position $\mathbf{x}$ through the spherical harmonics model:
\begin{equation}
    \mathbf{c} = \sum_{l=0}^{L} \sum_{m=-l}^{l} \mathbf{c}_{lm} Y_{lm}(\mathbf{d}(\bm{\mu}_x(t))),
\end{equation}
where $\mathbf{c}$ is the color of Gaussian primitive, $L$ is the degree of spherical harmonics, 
$\mathbf{c}_{lm}$ is the spherical harmonics coefficients, $\mathbf{d}(\bm{\mu}_x(t))$ is the view direction at position $\bm{\mu}_x(t)$,
and $Y_{lm}(\mathbf{d}(\bm{\mu}_x(t)))$ is the spherical harmonics basis function with direction $\mathbf{d}(\bm{\mu}_x(t))$.

The opacity of the moved Gaussian primitive at position $\mathbf{x}$ and time $t$ is defined as:
\begin{equation}
    \label{eq:opacity}
    \scalebox{0.8}{$
    \sigma(\mathbf{x}, t) = \sigma(t) * \sigma * \exp\left(-\frac{1}{2} \left(\mathbf{x}-\bm{\mu}_x(t)\right)^T \mathbf{\Sigma}^{-1} \left(\mathbf{x}-\bm{\mu}_x(t)\right)\right),
    $}
\end{equation}
where $\sigma$ is the original opacity of Gaussian primitive.
$\mathbf{\Sigma}$ is the covariance matrix defined by the scale and orientation of Gaussian primitive: 
$\mathbf{\Sigma}=RS{S}^{T}{R}^{T}$.
$\sigma(t)$ is the temporal opacity that aims to control the impact of the Gaussian primitive over time.
To enable the time and duration of Gaussian primitives to be automatically adjusted by rendering gradients, the temporal opacity should be a unimodal function with a scaling parameter.
Therefore, our approach models $\sigma(t)$ as a Gaussian distribution:
\begin{equation}
    \label{eq:temporal_opacity}
    \sigma(t) = \exp\left(-\frac{1}{2} \left(\frac{  t-\mu_t }{s}\right)^{2} \right),
\end{equation}
where $\mu_t$ and $s$ is the time and the duration of the Gaussian primitive.

\subsection{Training}
\label{sec:optimization_strategies}

Similar to 3DGS~\cite{kerbl20233d}, our approach optimizes the parameters of Gaussian primitives by minimizing the rendering loss between the observed images and the rendered images:
\begin{equation}
    \label{eq:rendering_loss}
    \mathcal{L}_{render} = \lambda_{img} \mathcal{L}_{img} + \lambda_{ssim} \mathcal{L}_{ssim} + \lambda_{perc} \mathcal{L}_{perc},
\end{equation}
where $\mathcal{L}_{img}$, $\mathcal{L}_{ssim}$, and $\mathcal{L}_{perc}$ are the image loss, SSIM loss~\cite{wang2004image}, and perceptual loss~\cite{zhang2018unreasonable}, respectively.

However, we find that simply optimizing the proposed representation with only rendering loss often leads to poor rendering quality in fast-moving or complex motion regions.
To address this problem, we analyze the distribution of Gaussian primitives' opacity and find that a significant portion of it is near 1.
Therefore, a plausible reason for the poor quality is that high opacity of some Gaussian primitives can prevent the gradient from backpropagating to all Gaussian primitives, hindering the optimization process.

\paragraph{4D regularization} Based on the observation, our approach designs a regularization loss to constrain high opacity values of Gaussian primitives, which is defined as:
\begin{equation}
    \label{eq:regularization}
    \mathcal{L}_{reg}(t) = \frac{1}{N} \sum_{i=1}^{N} \left(\sigma * sg\left[ \sigma(t) \right] \right),
\end{equation}
where $t$ is the time of observed images in each training iteration, $N$ is the number of Gaussian primitives, and $sg[\cdot]$ is the stop-gradient operation.
Here we introduce the temporal opacity $\sigma(t)$ as the weight of the regularization loss, which represents the impact of Gaussian primitives at a particular time.
For Gaussian primitives with less impact, we reduce the penalty on them.
Note that the stop-gradient operation is applied to prevent the regularization loss from minimizing the temporal opacity.

\paragraph{Periodic relocation of primitives}
Although the regularization loss can effectively improve the rendering quality, it causes a dramatic increase in the number of Gaussian primitives needed to represent the same scene.
To mitigate this, we design a periodic relocation strategy to move Gaussian primitives with low opacity to the region with high opacity.
Specifically, we design a sampling score $s$ for each Gaussian primitive to measure the region that requires more primitives:
\begin{equation}
    \label{eq:sampling_score}
    s = \lambda_{g} \triangledown_{g} + \lambda_{o} \sigma
\end{equation}
where $\triangledown_{g}$ and $\sigma$ are the spatial gradient and opacity of the Gaussian primitive, 
and $\lambda_{g}$ and $\lambda_{o}$ are the weights of the gradient and opacity, respectively.
For every $N$ iterations, we move the Gaussian primitives with opacity below a threshold to the region with high sampling score.

\paragraph{Initialization of our representation}
To further improve the rendering quality, our approach proposes a strategy to initialize the position, time, and velocity of Gaussian primitives.
For each video frame, we first use ROMA~\cite{edstedt2024roma} to obtain 2D matches across multi-view images and then calculate 3D points through 3D triangulation.
These 3D points and the corresponding time step are used to initialize the position and time of Gaussian primitives.
Subsequently, 3D points of two video frames are matched by k-nearest neighbor algorithm, and the translation between the point pairs are taken as the velocity of Gaussian primitives.

During the optimization process, we further anneal the optimization rate of velocity according to $\lambda_{t} = \lambda_{0}^{1-t} + \lambda_{1}^{t}$, where $t$ goes from 0 to 1 during training.
This annealing motion scheduler helps to model the fast motions in the early stage and the complex motion in the later stage.

\subsection{Implementation Details}

We implement our approach using PyTorch.
We use the Adam optimizer for optimization with the same settings as 3DGS.
The model is trained for 30k iterations for a sequence length of 300 frames, which takes around 1 hour on an RTX 4090 GPU.
The weight of the 4D regularization loss is $\lambda_{reg}$ set to $1e^{-2}$, 
and the weights of the image, SSIM and perceptual loss $\lambda_{img}, \lambda_{ssim}, \lambda_{perc}$  are set to 0.8, 0.2 and 0.01, respectively.
The weights of the gradient and opacity in the sampling score $\lambda_{g}, \lambda_{o}$ are set to 0.5 and 0.5, respectively.
And the periodic relocation is performed every $N=100$ iterations.

\section{Experiments}

\begin{figure*}[h]
    \centering
    \includegraphics[width=\linewidth]{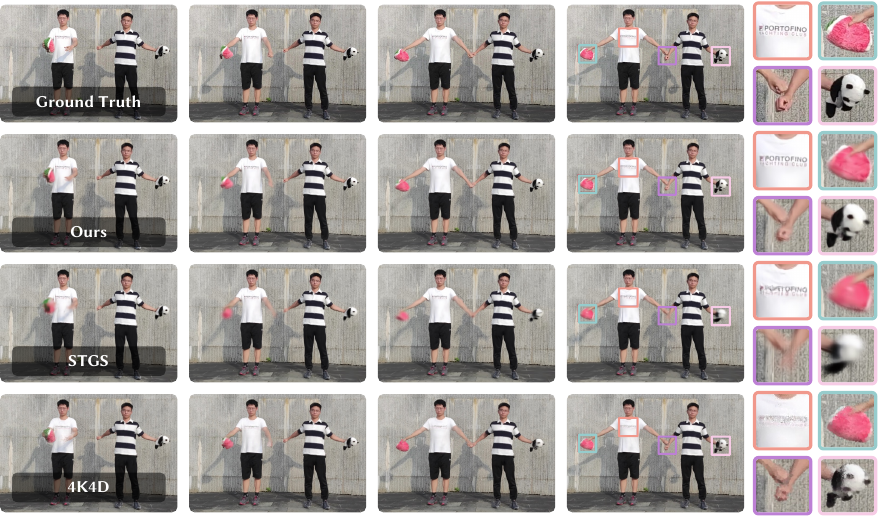}
    \caption{\textbf{Qualitative comparison on the ENeRF-Outdoor Dataset.}
    Our method achieves higher quality for fast-moving objects and regions, such as the swinging arms and dolls in hands, and clearer text details on the clothes.
    }
    \label{fig:compare_enerf_outdoor}
\end{figure*}

\begin{figure*}[h]
    \vspace{0.4cm}
    \centering
    \includegraphics[width=\linewidth]{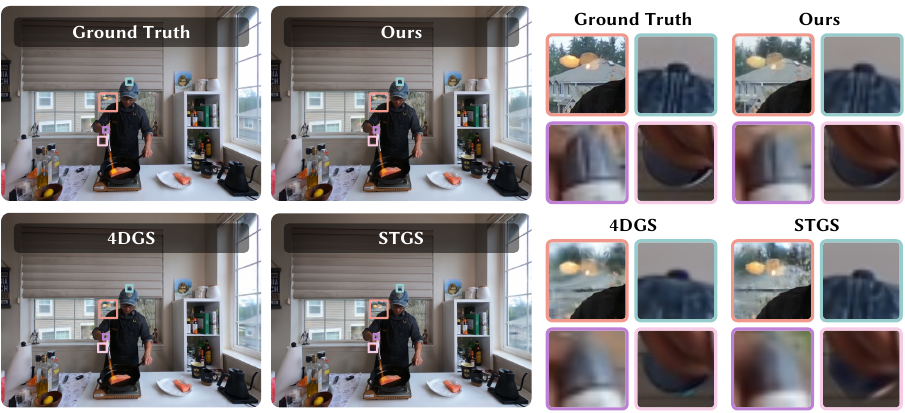}
    \caption{\textbf{Qualitative comparison on the Neural 3D Video Dataset.}
    Our method achieves the best rendering quality compared with baseline methods, especially for distant static regions and fast-moving dynamic regions.
    }
    \label{fig:compare_salmon}
\end{figure*}

\begin{figure*}[]
    \centering
    \includegraphics[width=\linewidth]{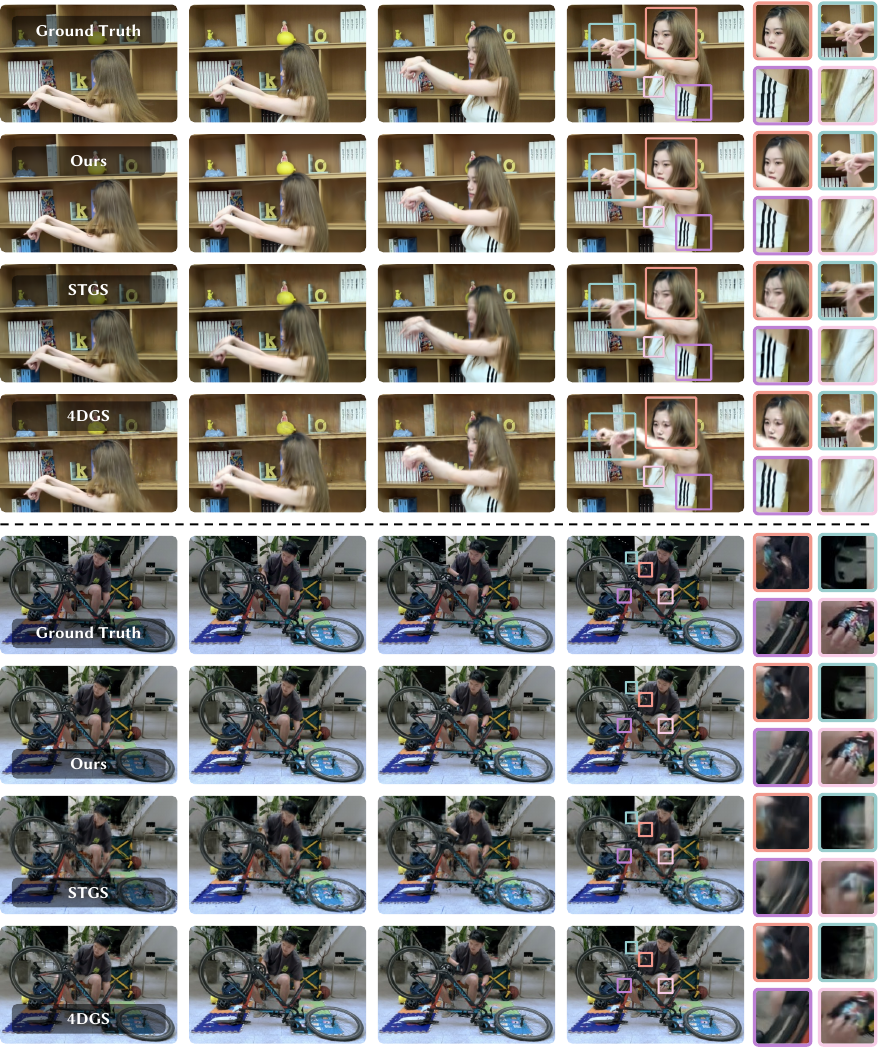}
    \caption{\textbf{Qualitative comparison on our \textit{SelfCap} Dataset.}
    Our method achieves significantly higher rendering quality than other methods. 
    For example, in the dance sequence, other methods struggle to handle fast-moving regions, such as fingers, faces, and texture details on clothes, while our method retains their details. 
    In the bike sequence, other methods failed to model the complex motions of hands and rapidly rotating pedals, while our method maintain high-quality rendering results.
    }
    \label{fig:compare_selfcap}
\end{figure*}

\begin{figure*}[h]
    \centering
    \vspace{-1 cm}
    \includegraphics[width=\linewidth]{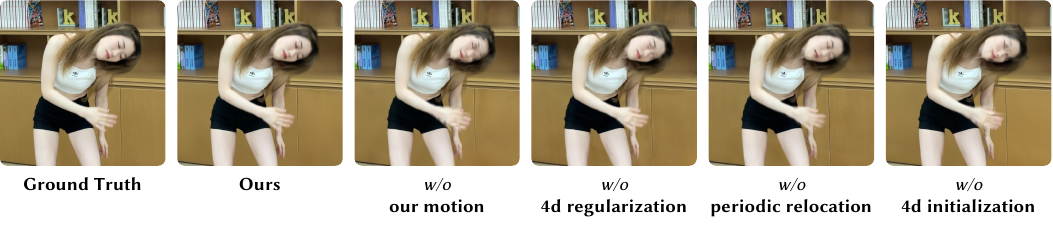}
    \vspace{-0.4 cm}
    \caption{\textbf{Ablation study of proposed components on dance1 sequence of our \textit{SelfCap} Dataset.}
    Removing our proposed components leads to visible artifacts in the rendered results, especially in the dynamic regions with fast motion.
    Our method produce high-quality results even in the challenging dynamic scenes.
    }
    \vspace{-0.3 cm}
    \label{fig:ablation_selfcap}
\end{figure*}

\paragraph{Datasets} 
We conducte our experiments using three datasets: Neural3DV~\cite{li2022neural}, ENeRF-Outdoor~\cite{lin2022enerf}, and our self-collected \textit{SelfCap} dataset. 
Neural3DV contains six scenes, each captured by 19-21 cameras with a resolution of 2704 $\times$ 2028 at 30 FPS. 
For each scene, we use the first 300 frames for training and evaluation, with an image resize ratio of 0.5.
ENeRF-Outdoor is a dynamic outdoor dataset with three scenes, captured by 18 synchronized cameras at a resolution of 1920 $\times$ 1080 and 60 FPS. 
We also use the first 300 frames for training and evaluation, with an image resize ratio of 1.
As the Neural3DV and ENeRF-Outdoor datasets lack significant motion scenes, we collect our own dataset, \textit{SelfCap}, to test performance in large-motion scenarios. 
\textit{SelfCap} contains eight scenes, each with 60 frames captured by 22-24 cameras, including daily life scenes such as dancing, playing with pets, and repairing bicycles.
Compared with existing datasets, \textit{SelfCap} contains more challenging scenes with fast and complex motion.
The resolution is 3840 $\times$ 2160 (1080 $\times$ 1080 for the \textit{bike} scene) at 60 FPS, with an image resize ratio of 0.5 (1 for the \textit{bike} scene).

\paragraph{Metrics} 
We use PSNR, DSSIM~\cite{wang2004image} and LPIPS~\cite{zhang2018unreasonable} to access the quality of the rendered images in our method and baselines. 
PSNR measures the $l_2$ difference between a reconstructed image and its ground truth, with higher values indicating less disparity. 
DSSIM$_1$ and DSSIM$_2$ measure the structural similarity, where higher values denote greater similarity. 
DSSIM$_1$ sets data range to $1.0$, while DSSIM$_2$ sets to $2.0$. 
LPIPS evaluates perceptual similarity, aligned with human perception. Lower LPIPS values indicate better perceptual similarity.

\subsection{Comparison Experiments}

\begin{table}[]
    \centering
    \caption{\textbf{Quantitative comparison on the Neural 3D Video~\cite{li2022neural} Dataset.} We report PSNR, DSSIM$_1$, DSSIM$_2$, and LPIPS  to evaluate the rendering quality. $^1$: only includes the \textit{Flame Salmon} scene. $^2$: excludes the \textit{Coffee Martini} scene.}
    \vspace{-0.2 cm}
    \resizebox{0.45\textwidth}{!}{
    \begin{tabular}{@{}lcccc@{}}
    \toprule
                         & PSNR$\uparrow$ & DSSIM$_1\downarrow$ & DSSIM$_2\downarrow$ & LPIPS$\downarrow$ \\ \midrule
    Neural Volume$^1$~\cite{Lombardi:2019}          & 22.80          & -                   & 0.062               & 0.295             \\
    LLFF$^1$~\cite{mildenhall2019llff}                   & 23.24          & -                   & 0.076               & 0.235             \\
    DyNeRF$^1$~\cite{li2022neural}                 & 29.58          & -                   & 0.020               & 0.083             \\
    HexPlane$^2$~\cite{cao2023hexplane}               & 31.71          & -                   & 0.014               & 0.075             \\
    K-Planes~\cite{fridovich2023k}                   & 31.63          & -                   & 0.018               & -                 \\
    MixVoxels-L~\cite{Wang2022MixedNV}                & 31.34          & -                   & 0.017               & 0.096             \\
    MixVoxels-X~\cite{Wang2022MixedNV}                & 31.73          & -                   & 0.015               & 0.064             \\
    HyperReel~\cite{attal2023hyperreel}                  & 31.10          & 0.036               & -                   & 0.096             \\
    NeRFPlayer~\cite{song2023nerfplayer}                 & 30.96          & 0.034               & -                   & 0.111             \\ \midrule
    Deformable-3DGS~\cite{wu20234d}            & 31.15          & 0.030               & -                   & 0.049             \\
    C-D3DGS~\cite{Katsumata2023ACD}                    & 30.46          & -                   & 0.022               & 0.150             \\
    SWinGS~\cite{Shaw2023SWinGSSW}                     & 31.10          & 0.030               & -                   & 0.096             \\
    Ex4DGS~\cite{Lee2024FullyED}                     & 32.11          & 0.030               & 0.015               & 0.048             \\
    4DGS~\cite{yang2023realtime}                       & 32.01          & -                   & 0.014               & 0.055             \\
    STGS~\cite{Li2023SpacetimeGF}                       & 32.05          & \textbf{0.026}               & 0.014               & 0.044             \\ 
    \midrule
    \textbf{Ours} & \textbf{33.19} & \textbf{0.026}      & \textbf{0.013}      & \textbf{0.036}     \\ \bottomrule
    \end{tabular}
    }
    \vspace{-0.2 cm}

    \label{tab:n3v}
\end{table}

\paragraph{Neural3DV} 
Qualitative and quantitative comparisons are shown in Figure~\ref{fig:compare_salmon} and Table~\ref{tab:n3v}, respectively. 
As evident in Table~\ref{tab:n3v}, our method outperforms all baselines in all metrics. 
Figure~\ref{fig:compare_salmon} illustrates our method’s ability to more accurately capture details in dynamic regions, such as the tail section of the flamethrower.
Additionally, our method can achieved better results in static background regions, exemplified by clearer rendering of the forest visible through the window.

\begin{table}[]
    \caption{\textbf{Quantitative comparison on the ENeRF-Outdoor~\cite{lin2022enerf} Dataset.} Green and yellow cell colors indicate the best and the second best results, respectively.}
    \centering\small
    \setlength{\tabcolsep}{9pt}
    \vspace{-0.1 cm}

    \begin{tabular}{lcccc}
    \toprule
                         & PSNR$\uparrow$ & DSSIM$_2\downarrow$ & LPIPS$\downarrow$ & FPS$\uparrow$ \\ \midrule
    ENeRF~\cite{lin2022enerf}                      & 24.96          & 0.107          & 0.299  & 3            \\
    4K4D~\cite{Xu20234K4DR4}                       & \cellsecond 25.28          & 0.096          & 0.379  & 220           \\
    4DGS~\cite{yang2023realtime}                       & 24.82          & \cellsecond 0.089          & 0.317  & 90           \\
    STGS~\cite{Li2023SpacetimeGF}                       & 24.93          & 0.091          & \cellsecond 0.297  & \cellsecond 226           \\ \midrule
    \textbf{Ours} & \cellfirst 25.36               & \cellfirst 0.077               & \cellfirst 0.244 & \cellfirst 454                 \\ \bottomrule
    \end{tabular}
    \vspace{-0.1 cm}

    \label{tab:enerf_outdoor}
\end{table}

\paragraph{ENeRF-Outdoor} 
Qualitative and quantitative comparisons are shown in Figure~\ref{fig:compare_enerf_outdoor} and Table~\ref{tab:enerf_outdoor}, respectively. 
This dataset introduces increased dynamic motion, as the actors exhibit extensive arm movements and manipulate a doll with a wide range of motion.
As indicated in Table~\ref{tab:enerf_outdoor}, our method achieves the highest performance across metrics.
Figure~\ref{fig:compare_enerf_outdoor} further demonstrates our method’s ability to capture fine details in rapid motion, particularly in objects such as the doll held by the actors and the text on the T-shirt.

\begin{table}[]
    \caption{\textbf{Quantitative comparison on our \textit{SelfCap} Dataset.} We include quantitative results for both the entire image and only dynamic regions (entire/dynamic). For 4DGS and STGS, we traversed different camera near plane settings during testing to maximize floater removal. Green and yellow cell colors indicate the best and the second best results, respectively.}
    
    \centering\small
    \setlength{\tabcolsep}{6pt}
    \vspace{-0.1 cm}

    \begin{tabular}{lcccc}
    \toprule
                            & PSNR$\uparrow$ & DSSIM$_2$$\downarrow$ & LPIPS$\downarrow$ & FPS$\uparrow$  \\ \midrule
    STGS~\cite{Li2023SpacetimeGF}               & 24.97/25.32          & 0.048/0.029          & 0.273/0.123 & \cellsecond 142                        \\
    4DGS~\cite{yang2023realtime}                    & \cellsecond 25.98/26.75          & \cellsecond 0.036/0.019          & \cellsecond 0.237/0.104 &  65                     \\
    \midrule
    Ours                    & \cellfirst 27.41/29.38               & \cellfirst 0.024/0.013               & \cellfirst 0.204/0.080 & \cellfirst 467                             \\ \bottomrule
    \end{tabular}
    \vspace{-0.1 cm}

    \label{tab:selfcap}
\end{table}

\paragraph{\textit{SelfCap}}
Qualitative and quantitative comparisons are shown in Figure~\ref{fig:compare_selfcap} and Table~\ref{tab:selfcap}, respectively.
Our method achieves the best performance on the \textit{SelfCap} dataset, which contains challenging dynamic scenes with fast and complex motion. 
As shown in Table~\ref{tab:selfcap}, our method outperforms all baselines in all metrics. In Figure~\ref{fig:compare_selfcap}, our method can better capture the details of the dynamic regions, like the fast-moving hands and the complex motion of the dancer's body.

\subsection{Ablation Studies}

\begin{table}[]
    \centering
    \caption{\textbf{Ablation studies.} We include quantitative results for both entire sequence and the subsequence with fastest motion (entire/fastest)}.
    \centering\small
    \setlength{\tabcolsep}{3pt}
    \vspace{-0.2 cm}

        \begin{tabular}{lccc}
            \toprule
            & PSNR$\uparrow$ & DSSIM$_2$$\downarrow$ & LPIPS$\downarrow$  \\ 
            \midrule
            w/o our motion   & 28.10/26.92          & 0.024/0.031          & 0.165/0.161                          \\
            w/o 4d regularization       & 28.68/29.09          & 0.023/0.024          & 0.159/0.150                          \\
            w/o periodic relocation     & \cellsecond 29.07/29.15          & \cellsecond 0.020/0.021          & \cellsecond 0.155/0.146                          \\
            w/o 4d initialization       & 28.33/27.06          & 0.023/0.030          & 0.162/0.158                          \\
            \midrule
            Ours                    & \cellfirst 29.74/30.75    & \cellfirst 0.018/0.017    & \cellfirst 0.152/0.133            \\
            \bottomrule
        \end{tabular}
    \label{tab:ablation}
    \vspace{-0.1 cm}

\end{table}

\begin{table}[]
  \caption{\textbf{Ablation studies.} We include quantitative results for both entire sequence and the subsequence with fastest motion (entire/fastest)}.
    \centering\small
    \setlength{\tabcolsep}{7pt}
    \vspace{-0.2 cm}

    \begin{tabular}{lccc}
    \toprule
      & PSNR$\uparrow$ & DSSIM$_2$$\downarrow$ & LPIPS$\downarrow$  \\ 
    \midrule
    $\lambda_{reg}=0$       &  28.68/29.09          & 0.023/0.024          & 0.159/0.150                          \\
    $\lambda_{reg}=1e^{-3}$     & \cellsecond 29.10/29.79          & \cellsecond 0.021/0.021          & \cellsecond 0.154/0.139                          \\
    $\lambda_{reg}=1e^{-2}$       & \cellfirst 29.74/30.75        & \cellfirst 0.018/0.017        & \cellfirst 0.152/0.133                          \\
    $\lambda_{reg}=1e^{-1}$       & 26.43/27.33        & 0.035/0.036        & 0.198/0.183                          \\
    \bottomrule
    \end{tabular}
    \vspace{-0.1 cm}

  \label{tab:ablation_reg}
\end{table}

We perform ablation studies on the 60-frame \textit{dance1} sequence of the \textit{SelfCap} dataset and further report the results on 10-frame with the fastest motion to evaluate the performance of our method in high-motion scenarios.
Quantitative and qualitative results are shown in Table~\ref{tab:ablation}, Table~\ref{tab:ablation_reg}, and Figure~\ref{fig:ablation_selfcap}.

\paragraph{Motion representation}
The "w/o our motion" variant removes the proposed motion representation and replace it with the motion representation used in 4DGS~\cite{yang2023realtime}.
As shown in Table~\ref{tab:ablation} and Figure~\ref{fig:ablation_selfcap}, our motion representation significantly improves the ability to model dynamic 3D scenes, especially in region with fast and complex motion.

\paragraph{4D regularization}
The "w/o 4d regularization" variant removes the 4D regularization loss $\mathcal{L}_{\text{reg}}$ in Eq.~\ref{eq:regularization}.
Without the 4D regularization, gaussians with large opacity will hinder the optimization process, leading to suboptimal results on detailed region.
We also ablate the effect of different $\lambda_{\text{reg}}$ values in Table~\ref{tab:ablation_reg} and choose $\lambda_{\text{reg}}=1e^{-2}$ for all of the experiments.

\paragraph{Periodic relocation}
The "w/o periodic relocation" variant use the same densify strategy as the 3DGS~\cite{kerbl20233d} and do not perform the proposed periodic relocation.
We also control the number of gaussians to be the same as the proposed method for a fair comparison.
Without the periodic relocation, the model tend to use more gaussians with lower opacity to model the scene, leading to suboptimal results when the number of gaussians is limited.

\paragraph{4D initialization}
The "w/o 4d initialization" variant removes the proposed velocity initialization method and use zero velocity initialization instead.
As shown in Table~\ref{tab:ablation} and Figure~\ref{fig:ablation_selfcap}, the proposed 4D initialization method significantly improve the model's ability to model the fast motion in the scene.

\section{Conclusions}

This paper introduces FreeTimeGS, a novel 4D representation method for dynamic 3D scenes. 
FreeTimeGS allows Gaussian primitives to emerge at any time and location, offering enhanced flexibility to model complex motions.
By assigning an optimizable explicit motion function and temporal opacity function to each Gaussian primitive, our representation can more faithfully and flexibly represent the dynamic scene.
Furthermore, we propose a simple regularization strategy that penalizes high opacity in Gaussian primitives, 
effectively mitigating the local minima problem during optimization. 
Experimental results demonstrate that FreeTimeGS achieves higher rendering quality and rendering speed on multiple widely-used datasets for multi-view dynamic novel view synthesis.

Notably, our method still has a few limitations. For one, our method still requires a length reconstruction process for each dynamic scene. Future work could potentially mitigate this by incorporating generative priors on the proposed representation for optimization-free reconstruction. Another limitation of our method is that the current representation doesn't support relighting, only focusing on novel view synthesis. Future work could extend the current representation with surface normal and material properties to extend its applicability for relighting.

\noindent{\bf{Acknowledgement}} This work was partially supported by NSFC (No. 62172364, No. U24B20154, No. 62402427), Zhejiang Provincial Natural Science Foundation of China (No. LR25F020003), and Information Technology Center and State Key Lab of CAD\&CG, Zhejiang University. We also acknowledge the EasyVolcap \cite{xu2023easyvolcap} codebase.
\clearpage
\appendix

\setcounter{page}{1}
\maketitlesupplementary

\section{More Experiments Results}
\subsection{More Quantitative Comparison on Neural3DV}
\label{sec:more_quantitative_results_neural3dv}
In Table~\ref{tab:n3v_sup}, we provide additional quantitative comparisons on the Neural3DV dataset including rendering speed and storage cost.

In Table~\ref{tab:ours_neural3dv_storage}, we further demonstrate the trade-off between rendering quality and storage cost by controlling the number of primitives of our method.

\subsection{More Quantitative Comparison on \textit{SelfCap}}
\label{sec:more_quantitative_results_selfcap}
In Table~\ref{tab:selfcap_sup}, we provide additional quantitative comparisons on the \textit{SelfCap} dataset including rendering speed and storage cost.
For 4DGS, we provide results with and without 4DSH. We find that 4DSH introduces additional storage cost without improving rendering quality on the \textit{SelfCap} dataset. Therefore, we do not use 4DSH in the \textit{SelfCap} experiments presented in the main paper.

\begin{figure*}[h]
    \centering
    \includegraphics[width=\linewidth]{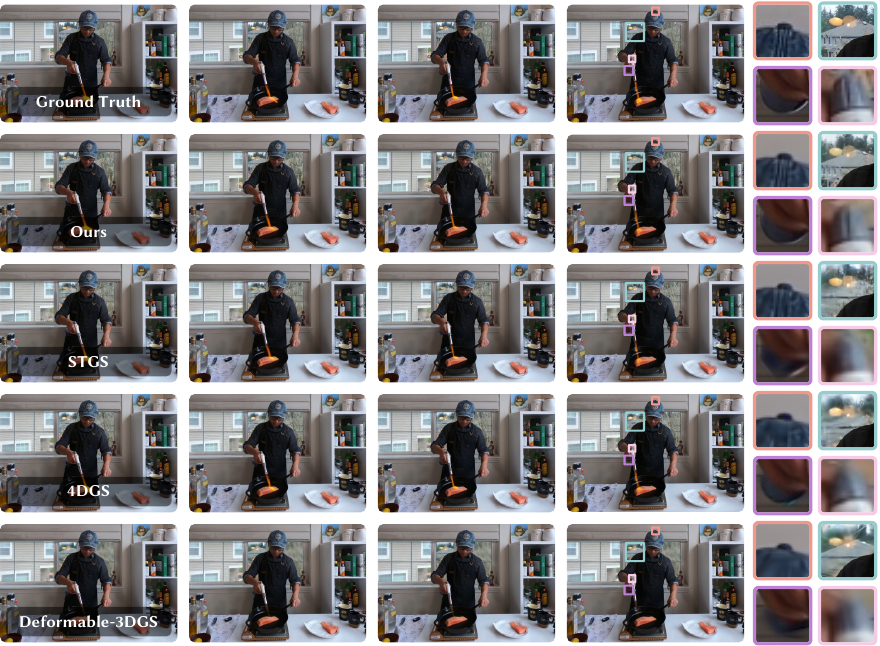}
    \caption{\textbf{Qualitative comparison on the Neural 3D Video Dataset.}
    }
    \label{fig:compare_salmon_full}
\end{figure*}

We also add the results of Deformable-3DGS~\cite{kerbl20233d} on the \textit{SelfCap} dataset. As stated in the main paper, the difficulty of building long-range correspondences makes it challenging to handle complex dynamic scenes, as shown in Table~\ref{tab:selfcap_sup}, Figure~\ref{fig:compare_selfcap_corgi_sup}, Figure~\ref{fig:compare_selfcap_dance_sup}, and Figure~\ref{fig:compare_salmon_full}.

\subsection{More Qualitative Results}
\label{sec:more_qualitative_results}
\begin{figure*}[]
    \centering
    \includegraphics[height=0.95\textheight]{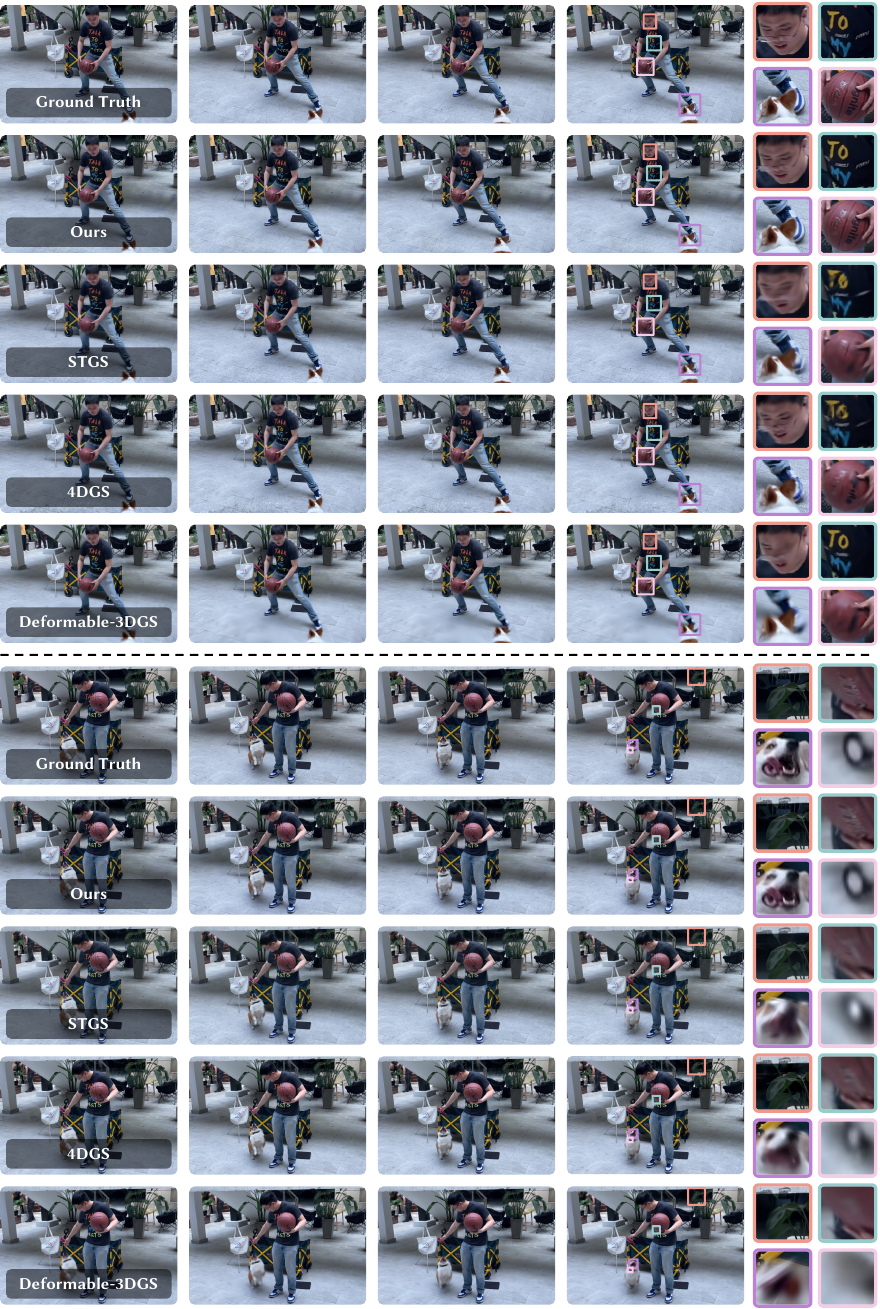}
    \caption{\textbf{Qualitative comparison on our \textit{SelfCap} Dataset.}
    }
    \label{fig:compare_selfcap_corgi_sup}
\end{figure*}

\begin{figure*}[]
    \centering
    \includegraphics[height=0.95\textheight]{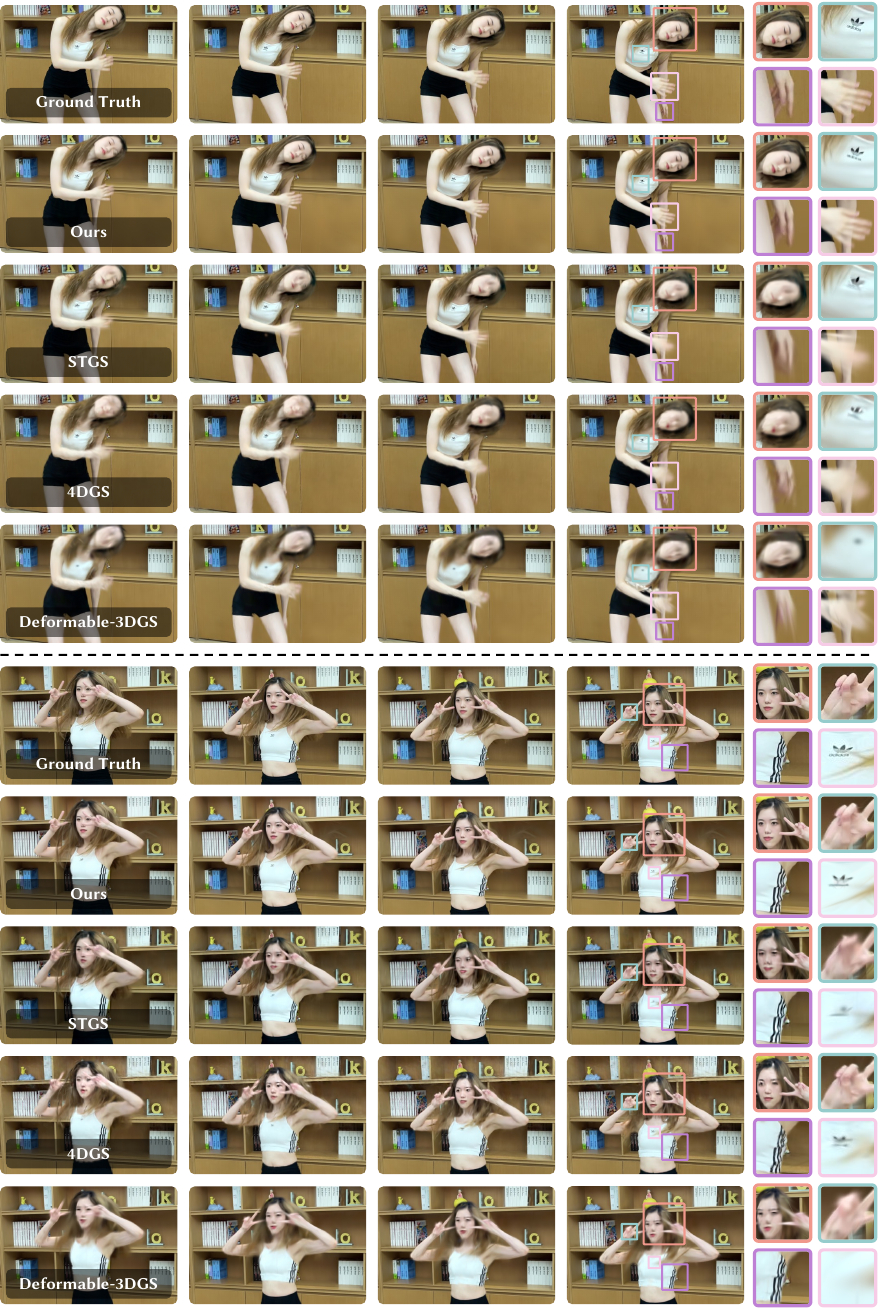}
    \caption{\textbf{Qualitative comparison on our \textit{SelfCap} Dataset.}
    }
    \label{fig:compare_selfcap_dance_sup}
\end{figure*}

More qualitative comparisons on the \textit{SelfCap} dataset are shown in Figure~\ref{fig:compare_selfcap_corgi_sup} and Figure~\ref{fig:compare_selfcap_dance_sup}.

\subsection{Per Scene Breakdown}
In Table~\ref{tab:n3v_perscene}, Table~\ref{tab:enerf_perscene}, and Table~\ref{tab:selfcap_perscene}, we provide the per-scene breakdown of the quantitative results on the Neural3DV, ENeRF-Outdoor, and \textit{SelfCap} datasets, respectively.
\begin{table}[]
    \centering
    \caption{\textbf{Quantitative comparison on the Neural 3D Video~\cite{li2022neural} Dataset.} We report PSNR, DSSIM$_1$, DSSIM$_2$, and LPIPS$_{Alex}$ to evaluate the rendering quality. $^1$: only includes the \textit{Flame Salmon} scene. $^2$: excludes the \textit{Coffee Martini} scene. $^\dag$: Our method with number of primitives controlled to no more than 500k.}
    \resizebox{0.45\textwidth}{!}{
    \begin{tabular}{@{}lccccc@{}}
    \toprule
                                            & PSNR$\uparrow$ & DSSIM$_1\downarrow$ & DSSIM$_2\downarrow$ & LPIPS$\downarrow$ & Size (MB) $\downarrow$ \\ \midrule
    Neural Volume$^1$~\cite{Lombardi:2019} & 22.80          & -                   & 0.062               & 0.295             & -                      \\
    LLFF$^1$~\cite{mildenhall2019llff}     & 23.24          & -                   & 0.076               & 0.235             & -                      \\
    DyNeRF$^1$~\cite{li2022neural}         & 29.58          & -                   & 0.020               & 0.083             & \textbf{28}                     \\
    HexPlane$^2$~\cite{cao2023hexplane}    & 31.71          & -                   & 0.014               & 0.075             & 200                    \\
    K-Planes~\cite{fridovich2023k}         & 31.63          & -                   & 0.018               & -                 & 311                    \\
    MixVoxels-L~\cite{Wang2022MixedNV}     & 31.34          & -                   & 0.017               & 0.096             & 500                    \\
    MixVoxels-X~\cite{Wang2022MixedNV}     & 31.73          & -                   & 0.015               & 0.064             & 500                    \\
    HyperReel~\cite{attal2023hyperreel}    & 31.10          & 0.036               & -                   & 0.096             & 360                    \\
    NeRFPlayer~\cite{song2023nerfplayer}   & 30.96          & 0.034               & -                   & 0.111             & 5130                   \\ \midrule
    Deformable-3DGS~\cite{wu20234d}        & 31.15          & 0.030               & -                   & 0.049             & 90                     \\
    C-D3DGS~\cite{Katsumata2023ACD}        & 30.46          & -                   & 0.022               & 0.150             & 338                    \\
    Ex4DGS~\cite{Lee2024FullyED}           & 32.11          & 0.030               & 0.015               & 0.048             & 115                    \\
    4DGS~\cite{yang2023realtime}           & 32.01          & -                   & 0.014               & 0.055             & 3128                   \\
    STGS-Lite~\cite{Li2023SpacetimeGF}     & 31.59          & 0.027               & 0.015               & 0.047             & 103                     \\
    STGS~\cite{Li2023SpacetimeGF}          & 32.05          & \textbf{0.026}      & 0.014               & 0.044             & 200                    \\ 
    \textbf{Ours}                          & \textbf{33.19} & \textbf{0.026}      & \textbf{0.013}      & \textbf{0.036}    & 125                       \\
    \textbf{Ours$^\dag$}                   & 32.97          & 0.028               & 0.014               & 0.043             & \textbf{41}                       \\ \bottomrule
    \end{tabular}
    }

    \label{tab:n3v_sup}
\end{table}

\begin{table}[]
  \caption{
    \textbf{Ablation studies on the number of primitives on Neural3D Video~\cite{li2022neural} Dataset.}We report PSNR, DSSIM$_1$, DSSIM$_2$, and LPIPS$_{Alex}$ to evaluate the rendering quality. 
    Our method achieve high-quality results even when the storage cost is reduced to as low as 8.3MB.
  }.
    \centering %

    \resizebox{0.45\textwidth}{!}{
    \begin{tabular}{@{}lccccc@{}}
    \toprule
    $\#$ Gaussians & PSNR$\uparrow$ & DSSIM$_1$$\downarrow$ & DSSIM$_2$$\downarrow$ & LPIPS$\downarrow$ & Size (MB) $\downarrow$ \\ \midrule
    $N =$ 70k        & 32.39          & 0.031                 & 0.016                 & 0.052             & \textbf{8.3}             \\
    $N =$ 91k        & 32.54          & 0.030                 & 0.015                 & 0.051             & 11                     \\
    $N =$ 165k       & 32.66          & 0.029                 & 0.015                 & 0.047             & 20                     \\
    $N =$ 347k       & 32.97 & 0.028        & 0.014        & 0.043    & 41                     \\ 
    $N =$ 618k       & 32.94 & 0.027        & 0.014        & 0.039    & 73                     \\ 
    $N =$ 1060k       & \textbf{33.19} & \textbf{0.026}        & \textbf{0.013}        & 0.036    & 125                     \\ 
    $N =$ 2042k       & 32.94 & 0.027        & \textbf{0.013}        & \textbf{0.035}    & 240                     \\ 
    
    \bottomrule
    \end{tabular}

    }
  \label{tab:ours_neural3dv_storage}
\end{table}

\begin{table}[]
    \caption{\textbf{Quantitative comparison on our \textit{SelfCap} Dataset.} We report PSNR, DSSIM$_2$, and LPIPS$_{VGG}$ to evaluate the rendering quality, and include quantitative results for both the entire image and only dynamic regions (entire/dynamic).
    The FPS is measured on an NVIDIA RTX 4090 GPU. Green and yellow cell colors indicate the best and the second best results, respectively.
    $^\dag$: Our method with number of primitives controlled to no more than 500k.}
    
    \centering

    \resizebox{0.45\textwidth}{!}{
    \begin{tabular}{lccccc}
    \toprule
                                                        & PSNR$\uparrow$                         & DSSIM$_2$$\downarrow$                  & LPIPS$\downarrow$                      & FPS$\uparrow$                  & Size (MB) $\downarrow$    \\ \midrule
    Deformable-3DGS~\cite{wu20234d}                     & 25.95/25.27                            & 0.037/0.026                            & 0.298/0.139                            & 57                             & \cellsecond 73                       \\
    STGS~\cite{Li2023SpacetimeGF} & 24.97/25.32                            & 0.048/0.029                            & 0.273/0.123                            & 142 & 77                       \\
    4DGS~\cite{yang2023realtime}  &  25.98/26.75 &  0.036/0.019 &  0.237/0.104 & 65                             & 827                      \\ 
    \textbf{Ours}                                                & \cellfirst 27.41/29.38  & \cellfirst 0.024/0.013  & \cellfirst 0.204/0.080  & \cellsecond 467  & 96                         \\
    \textbf{Ours$^\dag$}                                         & \cellsecond 27.27/28.87            & \cellsecond 0.025/0.013             & \cellsecond 0.217/0.081            & \cellfirst 664                               &  \cellfirst 53                        \\ \bottomrule
    \end{tabular}
    }

    \label{tab:selfcap_sup}
\end{table}

\begin{figure}[]
    \centering
    \includegraphics[width=1.0\linewidth]{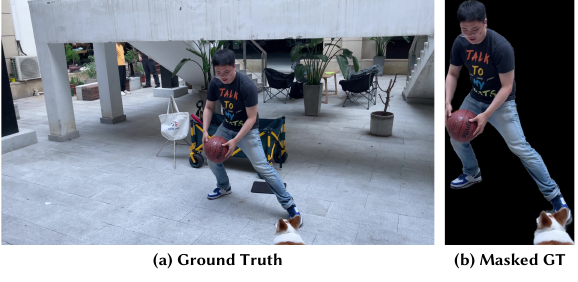}
    \caption{\textbf{Dynamic Region Mask.} An illustration of the use of dynamic region masking during evaluation. 
    (a) Ground Truth image from the \textit{SelfCap} dataset. (b) Masked Ground Truth, generated by cropping the Ground Truth image using a dynamic region mask extracted by Background Matting V2~\cite{Lin2020RealTimeHB}, with the outer area filled in black.
    }
    \label{fig:dynamic_region_mask}
\end{figure}

\section{Implementation Details}
\label{sec:implementation_details}
\subsection{Evaluation details on \textit{SelfCap}}
We report quantitative results for both the entire image and only dynamic regions on the \textit{SelfCap} dataset in the main paper and supplementary material.
For entire image evaluation, we use the full image to calculate the metrics.
For dynamic regions, we first use the ground truth background image and Background Matting V2~\cite{Lin2020RealTimeHB} to extract the mask of dynamic regions.
Then we crop the full image using the bounding box of the dynamic mask, and fill the region outside the mask with black color.
An example of the ground truth for dynamic region is shown in Figure~\ref{fig:dynamic_region_mask}.

\begin{table*}
	\centering
	\caption{\textbf{Quantitative comparison of view synthesis results on ENeRF-Outdoor~\cite{lin2022enerf} Dataset.}
    We report PSNR, SSIM$_2$, and LPIPS$_{VGG}$ to evaluate the rendering quality.
    Green and yellow cell colors indicate the best and the second best results, respectively.
    }
    \renewcommand{\arraystretch}{1.1}
    \resizebox{0.95\textwidth}{!}{
        \begin{tabular}{l|ccc|c|ccc|c|ccc|c}
            \Xhline{3\arrayrulewidth}
    & actor1\_4      & actor5\_6      & actor2\_3      & Avg.           & actor1\_4      & actor5\_6      & actor2\_3      & Avg.           & actor1\_4      & actor5\_6      & actor2\_3      & Avg.           \\ \cline{2-13} 
    & \multicolumn{4}{c|}{PSNR$\uparrow$}                               & \multicolumn{4}{c|}{SSIM$_2\uparrow$}                             & \multicolumn{4}{c}{LPIPS$\downarrow$}                             \\ \hline
ENeRF~\cite{lin2022enerf}         & \cellsecond{25.65} & 24.53          & 24.71          & 24.96          & 0.779          & 0.767          & 0.808          & 0.785          & \cellsecond{0.305} & 0.303          & \cellsecond{0.290} & 0.299          \\
4K4D~\cite{Xu20234K4DR4}          & \cellfirst{25.69} & 25.02          & \cellfirst{25.13} & \cellsecond{25.28} & 0.809          & 0.801          & 0.810          & 0.807          & 0.384          & 0.374          & 0.378          & 0.379          \\
4DGS~\cite{yang2023realtime}          & 24.78          & 24.77          & 24.91          & 24.82          & 0.782          & 0.785          & 0.794          & 0.787          & 0.326          & 0.320          & 0.306          & 0.317          \\
STGS~\cite{Li2023SpacetimeGF}          & 25.25          & \cellsecond{25.04} & 24.49          & 24.93          & \cellsecond{0.816} & \cellsecond{0.830} & \cellsecond{0.822} & \cellsecond{0.823} & 0.315          & \cellsecond{0.282} & 0.294          & \cellsecond{0.297} \\ \hline
\textbf{Ours} & 25.56          & \cellfirst{25.46} & \cellsecond{25.06} & \cellfirst{25.36} & \cellfirst{0.836} & \cellfirst{0.845} & \cellfirst{0.858} & \cellfirst{0.846} & \cellfirst{0.262} & \cellfirst{0.235} & \cellfirst{0.236} & \cellfirst{0.244} \\
\Xhline{3\arrayrulewidth}
\end{tabular}
    }

\label{tab:enerf_perscene}
\end{table*}

\newpage

\begin{table*}
	\centering
	\caption{\textbf{Quantitative comparison of view synthesis results on \textit{SelfCap} Dataset.}
    We report PSNR, SSIM$_2$, and LPIPS$_{VGG}$ to evaluate the rendering quality, and include quantitative results for both the entire image and only dynamic regions.
    Green and yellow cell colors indicate the best and the second best results, respectively.
    }
    \renewcommand{\arraystretch}{1.1}
    \resizebox{0.95\textwidth}{!}{
    \begin{tabular}{l|cccccccc|c}
    \Xhline{3\arrayrulewidth}
                    & dance\_1       & dance\_2       & dance\_3       & dance\_4       & corgi\_1       & corgi\_2       & bike\_1        & bike\_2        & Avg.           \\ \cline{2-10} 
                    & \multicolumn{9}{c}{Entire PSNR$\uparrow$}                                                                                                              \\ \hline
    Deformable-3DGS~\cite{wu20234d} & \cellsecond{26.82} & \cellsecond{26.54} & 25.69          & \cellsecond{26.80} & 24.15          & 28.04          & 24.30          & \cellsecond{25.25} & 25.95          \\
    4DGS-4DSH~\cite{yang2023realtime}            & 26.49          & 24.99          & 26.26          & 26.42          & 27.08          & 27.56          & \cellsecond{24.61} & 24.42          & \cellsecond{25.98} \\
    4DGS~\cite{yang2023realtime}       & 26.40          & 25.53          & \cellsecond{26.35} & 26.33          & 26.95          & 27.50          & 24.04          & 24.77          & \cellsecond{25.98} \\
    STGS~\cite{Li2023SpacetimeGF}            & 25.26          & 23.74          & 25.69          & 25.41          & \cellsecond{27.55} & \cellsecond{28.52} & 21.46          & 22.09          & 24.97          \\ \hline
    \textbf{Ours}   & \cellfirst{27.66} & \cellfirst{27.85} & \cellfirst{27.00} & \cellfirst{27.25} & \cellfirst{28.90} & \cellfirst{29.96} & \cellfirst{24.96} & \cellfirst{25.67} & \cellfirst{27.41} \\ \Xhline{3\arrayrulewidth}
                    & \multicolumn{9}{c}{Entire SSIM$_2\uparrow$}                                                                                                            \\ \hline
    Deformable-3DGS~\cite{wu20234d} & \cellsecond{0.940} & \cellsecond{0.936} & \cellsecond{0.943} & \cellsecond{0.940} & 0.900          & 0.902          & 0.920          & \cellsecond{0.924} & 0.926          \\
    4DGS-4DSH~\cite{yang2023realtime}            & 0.937          & 0.924          & 0.937          & 0.936          & 0.920          & 0.924          & \cellsecond{0.922} & 0.912          & \cellsecond{0.927} \\
    4DGS~\cite{yang2023realtime}       & 0.938          & 0.929          & 0.937          & 0.937          & 0.919          & 0.924          & 0.919          & 0.916          & \cellsecond{0.927} \\
    STGS~\cite{Li2023SpacetimeGF}            & 0.933          & 0.916          & 0.939          & 0.936          & \cellsecond{0.933} & \cellsecond{0.934} & 0.822          & 0.827          & 0.905          \\ \hline
    \textbf{Ours}   & \cellfirst{0.953} & \cellfirst{0.953} & \cellfirst{0.955} & \cellfirst{0.954} & \cellfirst{0.956} & \cellfirst{0.959} & \cellfirst{0.946} & \cellfirst{0.937} & \cellfirst{0.952} \\ \Xhline{3\arrayrulewidth}
                    & \multicolumn{9}{c}{Entire LPIPS$\downarrow$}                                                                                                           \\ \hline
    Deformable-3DGS~\cite{wu20234d} & 0.259          & 0.269          & 0.254          & 0.256          & 0.357          & 0.365          & 0.312          & 0.310          & 0.298          \\
    4DGS-4DSH~\cite{yang2023realtime}            & 0.246          & 0.269          & 0.221          & 0.225          & 0.212          & \cellsecond{0.210} & \cellsecond{0.256} & 0.275          & 0.239          \\
    4DGS~\cite{yang2023realtime}       & 0.244          & 0.257          & \cellsecond{0.219} & 0.223          & \cellsecond{0.211} & 0.212          & 0.259          & \cellsecond{0.269} & \cellsecond{0.237} \\
    STGS~\cite{Li2023SpacetimeGF}            & \cellsecond{0.237} & \cellsecond{0.245} & 0.222          & \cellfirst{0.217} & 0.219          & 0.235          & 0.406          & 0.402          & 0.273          \\ \hline
    \textbf{Ours}   & \cellfirst{0.234} & \cellfirst{0.232} & \cellfirst{0.213} & \cellsecond{0.218} & \cellfirst{0.158} & \cellfirst{0.160} & \cellfirst{0.199} & \cellfirst{0.221} & \cellfirst{0.204} \\ \Xhline{3\arrayrulewidth}
                    & \multicolumn{9}{c}{Dynamic PSNR$\uparrow$}                                                                                                             \\ \hline
    Deformable-3DGS~\cite{wu20234d} & 25.30          & 24.64          & 26.18          & 24.17          & 25.88          & 24.72          & 25.31          & \cellsecond{25.98} & 25.27          \\
    4DGS-4DSH~\cite{yang2023realtime}            & 27.10          & 22.17          & 27.47          & 25.84          & \cellsecond{29.32} & 27.52          & \cellsecond{26.39} & 25.58          & 26.42          \\
    4DGS~\cite{yang2023realtime}       & \cellsecond{27.26} & \cellsecond{26.27} & \cellsecond{27.54} & \cellsecond{26.08} & 26.62          & \cellsecond{27.72} & 26.05          & 26.45          & \cellsecond{26.75} \\
    STGS~\cite{Li2023SpacetimeGF}            & 26.95          & 25.00          & 27.05          & 25.63          & 28.54          & 26.81          & 21.15          & 21.41          & 25.32          \\ \hline
    \textbf{Ours}   & \cellfirst{29.74} & \cellfirst{28.91} & \cellfirst{29.24} & \cellfirst{27.33} & \cellfirst{31.20} & \cellfirst{32.62} & \cellfirst{28.89} & \cellfirst{27.07} & \cellfirst{29.38} \\ \Xhline{3\arrayrulewidth}
                    & \multicolumn{9}{c}{Dynamic SSIM$_2\uparrow$}                                                                                                           \\ \hline
    Deformable-3DGS~\cite{wu20234d} & 0.925          & 0.948          & 0.962          & 0.936          & 0.957          & 0.950          & 0.950          & 0.960          & 0.949          \\
    4DGS-4DSH~\cite{yang2023realtime}            & 0.945          & 0.938          & \cellsecond{0.968} & 0.955          & 0.976          & 0.972          & \cellsecond{0.957} & \cellsecond{0.962} & 0.959          \\
    4DGS~\cite{yang2023realtime}       & 0.945          & \cellsecond{0.959} & 0.966          & \cellsecond{0.957} & \cellsecond{0.977} & \cellsecond{0.973} & 0.956          & \cellsecond{0.962} & \cellsecond{0.962} \\
    STGS~\cite{Li2023SpacetimeGF}            & \cellsecond{0.949} & 0.944          & 0.966          & 0.956          & 0.967          & 0.966          & 0.890          & 0.896          & 0.942          \\ \hline
    \textbf{Ours}   & \cellfirst{0.963} & \cellfirst{0.970} & \cellfirst{0.975} & \cellfirst{0.965} & \cellfirst{0.982} & \cellfirst{0.990} & \cellfirst{0.975} & \cellfirst{0.967} & \cellfirst{0.973} \\ \Xhline{3\arrayrulewidth}
                    & \multicolumn{9}{c}{Dynamic LPIPS$\downarrow$}                                                                                                          \\ \hline
    Deformable-3DGS~\cite{wu20234d} & 0.232          & 0.155          & 0.138          & 0.182          & 0.113          & 0.112          & 0.102          & 0.075          & 0.139          \\
    4DGS-4DSH~\cite{yang2023realtime}            & 0.188          & 0.155          & 0.103          & 0.124          & 0.064          & 0.073          & \cellsecond{0.090} & 0.076          & 0.109          \\
    4DGS~\cite{yang2023realtime}       & 0.187          & 0.125          & 0.100          & 0.121          & \cellsecond{0.062} & \cellsecond{0.072} & \cellsecond{0.090} & \cellsecond{0.074} & \cellsecond{0.104} \\
    STGS~\cite{Li2023SpacetimeGF}            & \cellsecond{0.177} & \cellsecond{0.122} & \cellsecond{0.097} & \cellsecond{0.116} & 0.084          & 0.090          & 0.152          & 0.145          & 0.123          \\ \hline
    \textbf{Ours}   & \cellfirst{0.152} & \cellfirst{0.095} & \cellfirst{0.082} & \cellfirst{0.099} & \cellfirst{0.045} & \cellfirst{0.031} & \cellfirst{0.067} & \cellfirst{0.065} & \cellfirst{0.080} \\ \Xhline{3\arrayrulewidth}
    \end{tabular}
    }

\label{tab:selfcap_perscene}
\end{table*}

\begin{table*}
	\centering
	\caption{\textbf{Quantitative comparison of view synthesis results on Neural3D Video~\cite{li2022neural} Dataset. }
    We report PSNR, DSSIM$_1$, DSSIM$_2$, and LPIPS$_{Alex}$ to evaluate the rendering quality.
    Green and yellow cell colors indicate the best and the second best results, respectively.
    $^\dag$: Our method with number of primitives controlled to no more than 500k.
    }
    \renewcommand{\arraystretch}{1.1}
    \resizebox{0.95\textwidth}{!}{
    \begin{tabular}{l|cccccc|c}
    \Xhline{3\arrayrulewidth}
                    & Coffee Martini  & Cook Spinach    & Cut Roasted Beef & Flame Salmon    & Flame Steak     & Sear Steak      & Avg.           \\ \cline{2-8} 
                    & \multicolumn{7}{c}{PSNR$\uparrow$}                                                                                                    \\ \hline
    Neural Volume~\cite{Lombardi:2019}   & -               & -               & -                & 22.80           & -               & -               & 22.80          \\
    LLFF~\cite{mildenhall2019llff}            & -               & -               & -                & 23.24           & -               & -               & 23.24          \\
    DyNeRF~\cite{li2022neural}          & -               & -               & -                & 29.58           & -               & -               & 29.58          \\
    HexPlane~\cite{cao2023hexplane}        & -               & 32.04           & 32.55            & 29.47           & 32.08           & 32.39           & 31.71          \\
    K-Planes~\cite{fridovich2023k}        & 29.99           & 32.60           & 31.82            & 30.44           & 32.38           & 32.52           & 31.63          \\
    MixVoxels-L~\cite{Wang2022MixedNV}     & 29.63           & 32.25           & 32.40            & 29.81           & 31.83           & 32.10           & 31.34          \\
    MixVoxels-X~\cite{Wang2022MixedNV}     & 30.39           & 32.31           & 32.63            & 30.60           & 32.10           & 32.33           & 31.73          \\
    HyperReel~\cite{attal2023hyperreel}       & 28.37           & 32.30           & 32.92            & 28.26           & 32.20           & 32.57           & 31.10          \\
    NeRFPlayer~\cite{song2023nerfplayer}      & \cellfirst{31.53}  & 30.56           & 29.35            & \cellfirst{31.65}  & 31.93           & 29.13           & 30.69          \\
    Deformable-3DGS~\cite{wu20234d} & 27.34           & 32.46           & 32.90            & 29.20           & 32.51           & 32.49           & 31.15          \\
    Ex4DGS~\cite{Lee2024FullyED}          & 28.79           & 33.23           & 33.73            & 29.29           & 33.91           & 33.69           & 32.11          \\
    4DGS~\cite{yang2023realtime}            & 28.33           & 32.93           & 33.85            & 29.38           & 34.03           & 33.51           & 32.01          \\
    STGS~\cite{Li2023SpacetimeGF}            & 28.61           & 33.18           & 33.52            & 29.48           & 33.64           & 33.89           & 32.05          \\
    STGS-Lite~\cite{Li2023SpacetimeGF}       & 27.49           & 32.92           & 33.72            & 28.67           & 33.28           & 33.47           & 31.59          \\ \hline
    \textbf{Ours}            & \cellsecond{30.63}  & \cellfirst{33.80}  & \cellfirst{34.52}   & \cellsecond{31.18}  & \cellfirst{34.98}  & \cellsecond{34.06}  & \cellfirst{33.19} \\
    \textbf{Ours$^\dag$}           & 30.31           & \cellsecond{33.52}  & \cellsecond{34.13}   & 30.63           & \cellsecond{34.66}  & \cellfirst{34.56}  & \cellsecond{32.97} \\ \Xhline{3\arrayrulewidth}
                    & \multicolumn{7}{c}{DSSIM$_1\downarrow$}                                                                                                  \\ \hline
    HyperReel~\cite{attal2023hyperreel}       & 0.0540          & 0.0295          & 0.0275           & 0.0590          & 0.0255          & 0.0240          & 0.037          \\
    NeRFPlayer~\cite{song2023nerfplayer}      & \cellfirst{0.0245} & 0.0355          & 0.0460           & \cellfirst{0.0300} & 0.0250          & 0.0460          & 0.035          \\
    Deformable-3DGS~\cite{wu20234d} & 0.0475          & 0.0255          & 0.0215           & 0.0415          & 0.0230          & 0.0215          & 0.030          \\
    Ex4DGS~\cite{Lee2024FullyED}          & 0.0425          & 0.0265          & 0.0260           & 0.0415          & 0.0220          & 0.0205          & 0.030          \\
    STGS~\cite{Li2023SpacetimeGF}            & 0.0415          & \cellfirst{0.0215} & \cellfirst{0.0205}  & 0.0375          & \cellfirst{0.0176} & \cellfirst{0.0174} & \cellfirst{0.026} \\
    STGS-Lite~\cite{Li2023SpacetimeGF}       & 0.0437          & \cellsecond{0.0218} & \cellsecond{0.0209}  & 0.0387          & \cellsecond{0.0179} & \cellsecond{0.0177} & \cellsecond{0.027} \\ \hline
    \textbf{Ours}            & \cellsecond{0.0369} & 0.0233          & 0.0220           & \cellsecond{0.0347} & 0.0194          & 0.0198          & \cellfirst{0.026} \\
    \textbf{Ours$^\dag$}           & 0.0395          & 0.0242          & 0.0238           & 0.0369          & 0.0208          & 0.0199          & 0.028          \\ \Xhline{3\arrayrulewidth}
                    & \multicolumn{7}{c}{DSSIM$_2\downarrow$}                                                                                                  \\ \hline
    Neural Volume   & -               & -               & -                & 0.0620          & -               & -               & 0.062          \\
    LLFF            & -               & -               & -                & 0.0760          & -               & -               & 0.076          \\
    DyNeRF          & -               & -               & -                & 0.0200          & -               & -               & 0.020          \\
    K-Planes        & 0.0235          & 0.0170          & 0.0170           & 0.0235          & 0.0150          & 0.0130          & 0.018          \\
    MixVoxels-L     & 0.0244          & 0.0162          & 0.0157           & 0.0255          & 0.0144          & 0.0122          & 0.018          \\
    MixVoxels-X     & 0.0232          & 0.0160          & 0.0146           & 0.0233          & 0.0137          & 0.0121          & 0.017          \\
    Ex4DGS          & 0.0245          & 0.0120          & 0.0115           & 0.0220          & 0.0100          & 0.0105          & 0.015          \\
    STGS            & 0.0250          & \cellsecond{0.0113} & \cellsecond{0.0105}  & 0.0224          & \cellfirst{0.0087} & \cellfirst{0.0085} & \cellsecond{0.014} \\
    STGS-Lite       & 0.0270          & 0.0118          & 0.0112           & 0.0244          & 0.0097          & 0.0095          & 0.016          \\ \hline
    \textbf{Ours}            & \cellfirst{0.0198} & \cellfirst{0.0112} & \cellfirst{0.0101}  & \cellfirst{0.0186} & \cellfirst{0.0087} & 0.0091          & \cellfirst{0.013} \\
    \textbf{Ours$^\dag$}           & \cellsecond{0.0214} & 0.0117          & 0.0110           & \cellsecond{0.0197} & \cellsecond{0.0095} & \cellsecond{0.0089} & \cellsecond{0.014} \\ \Xhline{3\arrayrulewidth}
                    & \multicolumn{7}{c}{LPIPS$\downarrow$}                                                                                                   \\ \hline
    Neural Volume   & -               & -               & -                & 0.295           & -               & -               & 0.295          \\
    LLFF            & -               & -               & -                & 0.235           & -               & -               & 0.235          \\
    DyNeRF          & -               & -               & -                & 0.083           & -               & -               & 0.083          \\
    HexPlane        & -               & 0.082           & 0.080            & 0.078           & 0.066           & 0.070           & 0.075          \\
    MixVoxels-L     & 0.106           & 0.099           & 0.088            & 0.116           & 0.088           & 0.080           & 0.096          \\
    MixVoxels-X     & 0.081           & 0.062           & 0.057            & 0.078           & 0.051           & 0.053           & 0.064          \\
    HyperReel       & 0.127           & 0.089           & 0.084            & 0.136           & 0.078           & 0.077           & 0.099          \\
    NeRFPlayer      & 0.085           & 0.113           & 0.144            & 0.098           & 0.088           & 0.138           & 0.111          \\
    Ex4DGS          & 0.070           & 0.042           & 0.040            & 0.066           & 0.034           & 0.035           & 0.048          \\
    STGS            & 0.069           & 0.037           & \cellsecond{0.036}   & 0.063           & \cellsecond{0.029}  & \cellsecond{0.030}  & 0.044          \\
    STGS-Lite       & 0.075           & 0.038           & 0.038            & 0.068           & 0.031           & 0.031           & 0.047          \\ \hline
    \textbf{Ours}            & \cellfirst{0.052}  & \cellfirst{0.031}  & \cellfirst{0.030}   & \cellfirst{0.050}  & \cellfirst{0.026}  & \cellfirst{0.026}  & \cellfirst{0.036} \\
    \textbf{Ours$^\dag$}           & \cellsecond{0.065}  & \cellsecond{0.036}  & 0.037            & \cellsecond{0.060}  & 0.030           & 0.029           & \cellsecond{0.043} \\ \Xhline{3\arrayrulewidth}
    \end{tabular}
    }

\label{tab:n3v_perscene}
\end{table*}

{
    \small
    \bibliographystyle{ieeenat_fullname}
    \bibliography{main}
}

\end{document}